\documentclass{llncs}
\usepackage[vlined,ruled,linesnumbered]{algorithm2e}
\usepackage{multirow}
\usepackage{floatflt}
\usepackage{amsfonts}
\usepackage{bbold}

\usepackage[pdftex]{graphicx}
\graphicspath{{./fig/}}
\usepackage[cmex10]{amsmath}
\usepackage{url}
\usepackage{lscape}
\usepackage{array}
\newcolumntype{H}{>{\setbox0=\hbox\bgroup}c<{\egroup}@{}}

\begin{document}
%
%
\pagestyle{headings}  
%
\title{Should we Reload\\Time Series Classification\\ Performance Evaluation ?\\
(a position paper)}
\titlerunning{Time Series Classification Performance Evaluation}  
%
\author{Dominique Gay\inst{1} \and Vincent Lemaire\inst{2}}
\authorrunning{D. Gay\ V. Lemaire} 
\institute{Laboratoire d'Informatique et de Math\'ematiques EA2525\\
Universit\'e de La R\'eunion, La R\'eunion, France\\
\and
Orange Labs\\
Lannion, France}

\maketitle              

\begin{abstract}
Since the introduction and the public availability of the \textsc{ucr} time series benchmark data sets, numerous Time Series Classification (TSC) methods has been designed, evaluated and compared to each others. We suggest a critical view of TSC performance evaluation protocols put in place in recent TSC literature. The main goal of this ``position'' paper is to stimulate discussion and reflexion about performance evaluation in TSC literature.
\end{abstract}
%
\section{Introduction}
%
The \emph{need for time series data mining benchmarks}~\cite{KK03} has been fulfilled: with, firstly, the \textsc{ucr} Time Series Classification Archive~\cite{UCRArchive} and then the \textsc{uea} \& \textsc{ucr} Time Series Classification Repository~\cite{UEAArchive}, the research community now have more than 85 data sets from various application domains to evaluate newly introduced TSC methods and to compare them to existing ones. The public availability and the wide diffusion of the benchmark data had a strong and positive impact on the research community which has been prolific in terms of publications of TSC methods; as an example, the recent experimental evaluation in ~\cite{BLB+17} involves 18 recently proposed algorithms while in the same year, in 2017, two contenders, \textsc{BoPF}~\cite{LL17} and \textsc{weasel}~\cite{SL17} have been presented in top data mining conferences.

In most of the research papers, the experimental evaluation section starts with \emph{``Each UCR dataset provides a train and test split set which we use unchanged to make our results comparable the prior publications''}\footnote{quoted from \textsc{weasel} paper~\cite{SL17}.}. And this is where the reflexion we suggest begins. In the following, we will consider accuracy as the measure of predictive performance since it is the most widely used in TSC literature.


\section{Discussion \& reflexion}
\noindent \textbf{About train/test experiment.}\\
While in transactional data classification, resampling strategies (mainly bootstrap and stratified $k$-fold cross-validation) are the norm to estimate the expected performance of classifiers and to compare them~\cite{Sal97,DCB+14}, there is a singularity concerning predictive performance evaluation in TSC literature: the vast majority of research work restrains predictive performance evaluation to a single train/test split experiment~\cite{UEAArchive}, also called hold-out method or test sample estimation~\cite{Koh95}.

Unless disposing of a \emph{large and representative} data set of the application domain, a single train/test experiment is generally ineffective in providing predictive performance estimation and valuable comparison between TS classifiers without random subsamplings (i.e., repeated hold out experiments). Indeed, different samplings may lead to results with strong variations~\cite{Koh95}. Thus, a classifier $\mathcal{C}_A$ may show better predictive performance than another classifier $\mathcal{C}_B$ just because of this particular split. And, in such cases, subsequent statistical tests based on single train/test accuracy results (such as now commonly used post-hoc Nemenyi test~\cite{D06}) do not help more in comparing classifiers performance over several data sets since other train/test splits could have led to different accuracy and mean rank results and thus potentially different conclusions.

Moreover, unless disposing of tens of thousands instances, it is common to keep more instances in the training set than in the test hold out, generally around class-stratified ratio of 2/3 for training against the rest i.e., 1/3 for testing. The TSC repository~\cite{UEAArchive} provides various predefined train/test split ratios, going from 1.6\% to 81\% of the whole data set for training set (notice that 35 of the 85 TSC data sets show a train/test split ratio below 34\%). In addition to small --if not very small-- training sets, the train/test splits are not always class-stratified; it results in very (too) few representative instances of some class labels (especially in multi-class problems). We are aware that, in some application domains, ``labelled data is expensive to collect''~\cite{BLB+17}, however, considering the whole available and \emph{already-labelled} data, splitting for such small non-stratified training sets is also questionable. Here are some singularities that arise from some data sets drawn from the TSC archive:
\begin{itemize}
\item \textsc{DiatomSizeReduction}: a 4-class problem, with a train/test split ratio about 5\% (16 instances for training, 306 for testing) and a class distribution (1,6,5,4) for training against (33,92,94,87) for testing. The class distribution is not respected from train to test set. There are relatively 73\% more instances of class $c_1$ in the test set than in the training set.
\item \textsc{SonyAIRobotSurface1}: a 2-class problem, with a train/test split ratio about 3.22\% (20 instances for training, 601 for testing) and a class distribution (6,14) for training against (343,258) for testing. The class distribution is not respected from train to test set. There are relatively 90\% more instances of class $c_1$ in the test set than in the training set; and 38\% less instances of class $c_2$ in the test set than in the training set.
\item \ldots several additional similar cases of data sets can be found in commonly used benchmark~\cite{UEAArchive} : either the number of training examples is very small w.r.t the whole available data, either there is class distribution change between training set and test set (sometimes both singularities arise).
\end{itemize}
Generally speaking, in addition to the weakness of single train/test experiments for performance evaluation, choosing to split such way, without class-stratification in small training sets, results in ``unnecessary difficult'' TSC problems: indeed, the obtained training set is not always representative of the whole available data set and as explained above, it could lead to class distribution change between training and test sets (also known as prior probability shift~\cite{MRA+12}).
In such environments, intuitively, simple 1-Nearest Neighbor lazy learners~\cite{WMD+13} and ensemble methods that embed several 1-NN classifiers (with various distance measures or on various data representations)~\cite{BDH+12,BLH+15,LB15} still obtain ``good'' accuracy results since it is still possible to find a nearest neighbor of a test instance in very few training instances of a minor class. However, eager classifiers based on empirically observed per-class frequencies (such as Na{\"\i}ve Bayes or Decision Trees) often fail in characterizing the minor classes with very few representative instances.

If the TSC archive~\cite{UEAArchive} offers a wide variety of data sets, the original train/test splits also involves hidden difficult well-known problems in the Machine Learning community: e.g., learning from few examples or in class distribution change environment. Unfortunately, averaging ranks over all data sets to produce critical difference diagram presents a risk of hiding the reasons why a particular classifier shows better performance than another. As an example~\footnote{We did not integrate recently introduced \textsc{hive-cote}~\cite{LTB16} accuracy results since they are available under 100 resamples protocol.}, in Table~\ref{tab:train-test-comparisons}, we report performance comparisons of 11 recent classifiers like in recent literature (single train/test split following by significance testing). We provide several statistical tests by integrating step by step data sets which involves smaller size of training set.

\begin{table}[htbp!]
\centering
\scriptsize
\begin{tabular}{|l|c|HHc|ccccccccccc|}
\hline
Min. size & \#DB & NB C & $q_{\alpha}$ & CD & WEASEL & DTWCV & DTW & BOSS & LS & TSBF & ST & EE & CoTE & SNB & BoPF\\
\hline
$\geq 1000$ & 10 & 11 & 3.21865 & \emph{4.77403} & \textbf{3.250} & 8.300 & 8.600 & \textbf{5.700} & \textbf{6.050} & \textbf{7.700} & \textbf{\underline{3.050}} & \textbf{6.250} & \textbf{4.100} & \textbf{3.400} & 9.600\\
$>500$ & 22 & 11 & 3.21865 & \emph{3.21865} & \textbf{\underline{2.932}} & 7.659 & 9.023 & \textbf{5.705} & 6.909 & 6.841 & \textbf{4.068} & 6.273 & \textbf{3.636} & \textbf{4.000} & 8.955\\
$>300$ & 42 & 11 & 3.21865 & \emph{2.32949} & \textbf{3.560} & 7.929 & 8.893 & 5.857 & 6.845 & 6.298 & \textbf{4.500} & 6.060 & \textbf{\underline{3.369}} & \textbf{4.821} & 7.869\\
$>200$ & 48 & 11 & 3.21865 & \emph{2.17903} & \textbf{3.594} & 7.917 & 8.885 & 5.781 & 6.885 & 6.271 & \textbf{4.625} & 5.865 & \textbf{\underline{3.219}} & \textbf{5.010} & 7.948\\
$>100$ & 57 & 11 & 3.21865 & \emph{1.99962} & \textbf{3.737} & 7.868 & 8.860 & 5.474 & 6.623 & 6.237 & \textbf{4.544} & 5.930 & \textbf{\underline{3.298}} & 5.535 & 7.895\\
All & 85 & 11 & 3.21865 & \emph{1.63748} & \textbf{3.847} & 7.806 & 8.647 & 5.382 & 6.135 & 6.388 & \textbf{4.847} & 5.871 & \textbf{\underline{3.412}} & 6.429 & 7.235\\
\hline
\end{tabular}
\caption{Average ranks of 11 recent classifiers depending on the data sets taken into account from~\cite{UEAArchive} (i.e., depending on the training set size). Accuracy results are taken from Sch{\"{a}}fer \& Leser paper on WEASEL~\cite{SL17}. Post-hoc Nemenyi's statistical test considering training set with size $\geq 1000$ (only 10 data sets), then $> 500$ (only 22 data sets), \ldots, until considering All 85 data sets from~\cite{UEAArchive}. Underlined rank is the best per line and bold results on the same line indicates that there is no statistically significant difference of performance between bold results according to Nemenyi's test, considering the current benchmark datasets.}
\label{tab:train-test-comparisons}
\end{table}

\textsc{weasel} and \textsc{ST}~\cite{HLB+14,BB15} score the highest mean ranks when considering data sets with training set size greater than 500 while \textsc{CoTE} takes advantage when adding data with smaller training size. We also observe that the mean rank of \textsc{ST} increases as we consider more and more data sets with smaller training size. Notice that, it has to be balanced against the fact that as the number of data sets decreases the critical difference (CD) value increases, making more difficult to state significant differences of performance between contenders. Another illustrative example is about \textsc{snb}~\cite{Bou14} (an improved Na{\"\i}ve Bayes classifier benefiting from multiple representations) which is competitive with \textsc{weasel}, \textsc{CoTE}, \textsc{boss} and \textsc{st} when not considering too small ($<200$) training set size -- confirming the importance of the size of the training set on the predictive performance of some classifiers.
Aside from~\cite{WMD+13}, we may regret the lack of experimental studies about the learning curves~\cite{PPS03,TWW+17} of TSC algorithms.\\


\noindent \textbf{About resampling strategies.}\\
As far as we know, very few attempts of resampling strategies for TSC performance evaluation have been led, e.g., :
\begin{itemize}
\item Grabocka et al.~\cite{GNS12} provides some results by 5-folds cross validation on 35 \textsc{ucr} data sets. It allows to identify easy data sets in the TSC archive \cite{UCRArchive}. Indeed, a default SVM classifier with polynomial kernel scores above 95\% accuracy (often near perfect) on 18 data sets.
\item Wang et al.~\cite{WMD+13}, focusing on 1-NN with various distance measures, provides $k$-folds stratified cross-validation results. However, the $k$ varies from 2 to 30 depending on the benchmark data set and the cross-validation method at use is unconventional: when splitting the data set $T$ into $k$ folds, the model is learnt on fold $T_k$ and tested on $T\setminus T_k$ --while conventional $k$-folds cross validation being the opposite: learning the model on $T\setminus T_k$ and testing on $T_k$. This unconventional cross-validation leads to the same problems explained above (with 1/30, i.e., 3\% of the whole data set used for training).\\
In the same paper, the authors also noticed the importance of the effect of the training data set size on 1-NN classifier accuracy : DTW-1-NN is better than ED-1-NN with small training data set, but providing that we have enough training instances (a few hundred/thousand depending on the simulated data set), the two classifiers show similar predictive performance.
\item Bagnall et al.~\cite{BLB+17} performs 100 resampling experiments on each of the 85 TSC data sets --followed by Nemenyi's statistical post-hoc test. However, the multiple resamplings fit the original size of train/test split provided by~\cite{UEAArchive} -- which leads to the same problem raised above about training size and class distribution change. This resampling strategy gives, all the same, a better idea of the performance of recent classifiers on data with \emph{various} sizes of training sets, although the train/test split is still questionable. Indeed, if instances from original test set are authorized to be in training set due to resampling, why not use 10-CV or 10$\times$10 CV, as in transactional data classification literature\footnote{However, even with $k$-folds CV, a particular attention must be given to the setting of $k$ for the 17 small data sets, with less than 200 instances, from the TSC archive.} ?
\end{itemize}
The cross-validation (CV) method is not unknown to the TSC community; indeed, some algorithms, like e.g., \textsc{weasel} or \textsc{DTW-CV} used cross-validation on training set to set hyper-parameters (even if the training set is very small), then a single train/test split is performed to evaluate the performance of the ``best hyper-parametered'' model on a single hold out test set. Again, why not use CV method to evaluate and compare TSC algorithms ?\\

\noindent{\textbf{About (repeated) 10-folds CV, statistical tests and beyond.}}\\
While 10-folds CV with subsequent statistical tests~\cite{D06} is now the gold standard for predictive performance evaluation and comparisons between classifiers on transactional benchmark data sets, recently, Vanwinckelen \& Blockeel~\cite{VB12,VB14} warns the Machine Learning community about pitfalls hidden in such comparisons. The take away messages are:
\begin{itemize}
\item \emph{``Our experiments show that when using cross-validation for choosing between two models, the best performing model is not always chosen''.}
\item \emph{``This discussion leads us to question the usefulness of statistical testing in the context of
evaluating predictive models with cross-validation''.}
\end{itemize}
On the other hand, after almost a decade of the \emph{``10-CV + statistical tests''} combination~\cite{D06} to evaluate learner's predictive performance, J. Demsar et al.~\cite{BCD+17,CBD+17} ``\emph{discourage the use of frequentist null hypothesis significance tests (NHST) in machine learning and, in particular, for comparison of the performance of classifiers}'' and encourages the community to embrace Bayesian analysis using 10$\times 10$-CV for comparing classifiers. This is perhaps a change point for performance evaluation habits in the Machine Learning community. Notice that Bayesian analysis of performance is more conservative than NHST; that is, it is ``more difficult'' for a classifier $\mathcal{C}_A$ to be better than a classifier $\mathcal{C}_B$, considering Bayesian analysis.\\


\noindent{\textbf{About the evaluation measure.}}\\
As recalled in the introduction, the vast majority of recently proposed TSC algorithms are evaluated and compared with regards to the accuracy measure, i.e., the number of correctly classified time series. Accuracy measure is suitable for roughly balanced 2-class data sets. However, for unbalanced and/or multiclass data sets, accuracy measure is inappropriate for evaluation since high accuracy results due to a bias towards the majority class could hide severely bad predictive performance on the minor class or on other classes in multiclass settings. Often, ROC or Precision/Recall curve analysis or lift curve and cumulative gain charts are prefered for unbalanced settings. 

The TSC archive~\cite{UEAArchive} contains some 2-class unbalanced problems, e.g., Earthquakes, ToeSegmentation2 and Wafer with respectively 20.2\%, 25.3\% and 10.6\% unbalanced ratio (i.e., the proportion of the minor class). The archive also contains many multiclass problems with severe unbalanced ratios, e.g., ECG5000 and Worms with respectively 0.5\% and 9.7\% unbalanced ratio. Learning in the presence of class imbalance or in multiclass settings is still an ongoing machine learning research topic~\cite{WMC+17}. Again, the presence of such data sets in the benchmark repository, when averaging ranks of classifiers based on accuracy results, could lead to flawed conclusions on the performance evaluation.\\

\section{Conclusion}
The still ongoing public release of benchmark TSC data sets to the data mining community through the \textsc{ucr \& uea} repository has certainly been the best catalyst for the development of new TSC algorithms and methods. In this discussion paper, we briefly review the habitual protocols at use in TSC algorithms performance evaluation, discuss the pros and cons of experimental protocols and try to warn the community about the pitfalls and hidden problems of actual performance evaluation protocols in TSC literature. We agree that the core of this discussion paper needs more in-depth development and experimental arguments but we believe that interesting discussions on this important topic deserve to be launched and continued during the $3^{rd}$ ECML/PKDD Workshop on Advanced Analytics and Learning on Temporal Data.\\

\section*{Acknowledgments}
We would like to thank the anonymous reviewers for joining the discussion with valuable comments and arguments.


\bibliographystyle{splncs03}
\bibliography{biblio_general}

\end{document}